\documentclass[letterpaper, 10 pt, conference]{ieeeconf}  

\IEEEoverridecommandlockouts                              

\usepackage[utf8]{inputenc}
\usepackage[T1]{fontenc}
\usepackage{graphicx}
\usepackage{multicol}
\usepackage{dblfloatfix}
\usepackage{epsfig} 
\usepackage{amsmath} 
\usepackage{amssymb}  
\usepackage{subfig}
\usepackage{float}
\usepackage{hyperref}
\usepackage{cleveref}

\usepackage{color}

\usepackage[sc]{mathpazo}
\usepackage[T1]{fontenc}

\usepackage[font=footnotesize,labelfont=bf]{caption}
\usepackage{cite}
\usepackage{mathrsfs}
\usepackage{algorithm, algorithmic}
\renewcommand{\baselinestretch}{1}
\newcommand{\R}{\mathcal{R}}
\newcommand{\Hu}{\mathcal{H}}

\title{\LARGE \bf
	Behavior Planning of Autonomous Cars with Social Perception
}

\author{Liting Sun$^{1}$, Wei Zhan$^{1}$, Ching-Yao Chan$^{2}$ and Masayoshi Tomizuka$^{1}$
	\thanks{*This work was supported by Mines ParisTech Foundation, ``Automated Vehicles - Drive for All''. }
	\thanks{$^{1}$Liting Sun, Wei Zhan and Masayoshi Tomizuka are with the Department of Mechanical Engineering, University of California at Berkeley, CA, 94720, USA.
		{\tt\small \{litingsun, wzhan, tomizuka\}@berkeley.edu}}%
	\thanks{$^{2}$Ching-Yao Chan is with California PATH, University of California at Berkeley, CA, 94720, USA.
		{\tt\small cychan@berkeley.edu}}%
}

\begin{document}

	\maketitle
	\thispagestyle{empty}
	\pagestyle{empty}

	\begin{abstract}
		Autonomous cars have to navigate in dynamic environment which can be full of uncertainties. The uncertainties can come either from sensor limitations such as occlusions and limited sensor range, or from probabilistic prediction of other road participants, or from unknown social behavior in a new area. To safely and efficiently drive in the presence of these uncertainties, the decision-making and planning modules of autonomous cars should intelligently utilize all available information and appropriately tackle the uncertainties so that proper driving strategies can be generated. In this paper, we propose a social perception scheme which treats all road participants as distributed sensors in a sensor network. By observing the individual behaviors as well as the group behaviors, uncertainties of the three types can be updated uniformly in a belief space. The updated beliefs from the social perception are then explicitly incorporated into a probabilistic planning framework based on Model Predictive Control (MPC). The cost function of the MPC is learned via inverse reinforcement learning (IRL). Such an integrated probabilistic planning module with socially enhanced perception enables the autonomous vehicles to generate behaviors which are defensive but not overly conservative, and socially compatible. The effectiveness of the proposed framework is verified in simulation on an representative scenario with sensor occlusions.
	\end{abstract}

	\section{INTRODUCTION}
	\label{sec:intro}
	The driving environment of autonomous vehicles (AVs) are dynamic and can be full of uncertainties. First, the future behaviors and trajectories of other traffic participants, such as pedestrians or vehicles with human drivers, are probabilistic in nature. It is difficult to predict them precisely, particularly in highly interactive driving scenarios. Beyond that, the implicit social behavior on local driving preferences and styles is also hard to describe exactly when the AVs are adapting themselves to a new environment. Moreover, the detection and tracking modules can produce lots of physical state uncertainties due to the algorithmic limitation in terms of unsatisfactory performance, as well as the physical limitation such as sensor field-of-view occlusions and limited sensor range.
	
	To generate safe and efficient maneuvers of autonomous vehicles, the decision-making and planning modules of AVs should be able to properly tackle all the uncertainties in the preceding modules such as perception and prediction. Research efforts were devoted recently to designing decision-making and planning algorithms under behavioral uncertainties from prediction. For example, an interactive belief-state planner proposed in \cite{hubmann_belief_2018} used Partially Observable Markov Decision Process (POMDP) to deal with the behavior uncertainties of other vehicles. A decision-making framework was also constructed in \cite{noh_decision-making_2019} to deal with uncertain behavior of other vehicles at intersections considering potential violations.
	
	\begin{figure}[t!]
		\begin{center}
			\includegraphics[width=8.5cm]{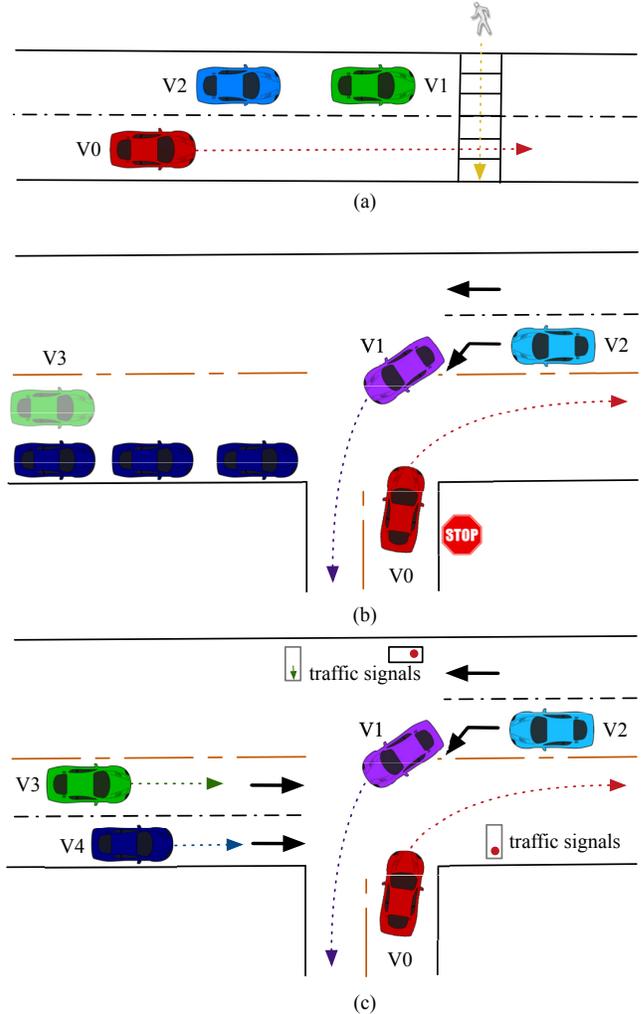}
			\caption{Exemplar scenarios where other road users can serve as distributed social sensors for the host vehicle (red).}
			\label{fig: exemscene}
		\end{center}
	\end{figure}
	
	While focusing on the behavior uncertainties, there is a common assumption in the aforementioned work, that is, the physical states of other traffic participants are deterministic and accurate. The perception module is assumed to provide perfect results on whether an object exist or not, or what the current positions, velocities and orientations of different objects are. However, such an assumption can hardly hold in practice. Even if the sensors, such as cameras and LiDARs, can well capture the objects, it is impossible for the most state-of-the-art algorithms to achieve perfect perception, as evidenced by the 3D detection results for the "easy" cases on KITTI benchmark \cite{kitti}.
	
	In addition to the algorithmic limitations, the physical limitations can also lead to uncertainties of the physical states of objects. Physical limitations in perception mainly include occlusion and limited sensor range \cite{orzechowski_tackling_2018}. Occlusion, an inevitable encounter of autonomous vehicles, causes great challenges for tracking, prediction and risk assessment \cite{galceran_augmented_2015} \cite{li_generic_2018}\cite{yu_occlusion-aware_2018}, and therefore poses significant impacts on the performance of decision-making and planning. To deal with occlusions, a safe driving strategy was proposed in \cite{hoermann_entering_2017} at blind intersections. Focusing also on blind intersections, \cite{morales_towards_2018} directly designed a planning method based on inverse reinforcement learning (IRL). 
	In \cite{bouton_scalable_2018}, a decision-making approach under occlusions was proposed in the framework of POMDP. 
	
	Most of the works, however, treats all other traffic participants such as pedestrians or human-driven vehicles only as objects/obstacles to avoid. In fact, they are all intelligent agents whose behaviors can be quite informative. \textit{Hence, our key observation is that human participants should be treated not only as dynamic obstacles, but also as distributed sensors that provide via their behaviors additional information about the environment beyond the scope of physical sensors. We call this concept \textbf{social perception}. The decision-making and planning modules of autonomous cars should explicitly exploit the enhancement offered by the social perception.}
	
	Figure \ref{fig: exemscene} demonstrates several exemplar scenarios where other road users can serve as sensors to overcome occlusions or limited sensor range. In Fig.~\ref{fig: exemscene}(a), the host vehicle V0 cannot detect the pedestrian due to the occlusion caused by V1 and V2. However, it can be inferred that the most probable reason for V1 to decelerate is a pedestrian crossing the street. Therefore, the behavior of V1 can be exploited as a sensor to enable social perception for potential pedestrians. In Fig.~\ref{fig: exemscene}(b), the host vehicle V0 is making a right turn at a one-way-stop T-intersection. It should yield to a potential vehicle V3, but the view is occluded by street-parked vehicles. However, V1 and V2 on the left-turn-only lane keep moving and making left turn, which indicates that there should be no vehicle in the occluded area or a vehicle if any might be relatively far away, and V0 may proceed to turn. Figure \ref{fig: exemscene}(c) shows a signalled intersection. The host vehicle V0 (turning right) can only detect the signal (red light) in front of it controlling its direction. It should yield to V3 and V4 which are still with relatively high speeds. However, V1 and V2 on the left-turn-only lane accelerate, which indicates that there is a protected left turn for them, and V0 can proceed to turn right. Therefore, the social perception is needed when the motion attributes of others are out of the limited sensor range.
	
	Inferring the physical states from the behavior of others as described in the examples is one aspect of social perception. For example, \cite{afolabi_people_2018} infers the map occupancy from behaviors of human drivers based on manually designed rules.  A more important aspect for social perception is that it can go beyond the perception of physical states and extend to the perception of social information existed within a group of social agents. Courtesy \cite{sun_courteous_2018} is one of the representative social information to be extracted. Socially cohesive behavior was analyzed and designed in \cite{landolfi_social_2018} by assuming that the behaviors of others (for instance, human drivers) were often correct and similar behaviors should be generated by autonomous vehicles. 
	
	Integrating the social perception into the decision-making and planning modules of autonomous vehicles is extremely important to enable safer and more efficient maneuvers in the presence of corresponding uncertainties. Collisions could be potentially avoided (Fig.~\ref{fig: exemscene}(a)) and the behavior of the autonomous vehicle can be more efficient, less conservative (Fig. \ref{fig: exemscene}(b) and (c)),  and more socially compatible so that both the passengers and the other human drivers will not be surprised or annoyed. In this paper, we explicitly incorporate the social perception into a probabilistic planner based on Model Predictive Control (MPC), and propose a unified planning framework to handle the above mentioned uncertainties.
	
\section{PROBLEM STATEMENT}
	\label{sec:PS}
	In this paper, we consider the behavior planning of an autonomous car in a multi-agent environment with perception uncertainties. Except for the autonomous car, denoted as $\R$, we assume all other agents to be human, represented by $\Hu$. Hence, we do not explicitly model the interactions among human, but focus on the interaction between the robot car and an individual human. 
   As for the perception uncertainties, we consider two types of uncertainties as defined above: the physical state uncertainties such as occlusions and limited sensor range, and the social behavioral uncertainties such as local driving preferences. 

	Throughout the paper, we let $x_{\R}$ and $u_{\R}$ denote, respectively, the robot car's states and control inputs, and $x_{\Hu, i}$ and $u_{\Hu, i}$ for those of human $i$. In a traffic scene with $M$ human participants, the states of all agents become  $x_{A}{=}(x_{\R}^T,x_{\Hu, 1}^T, \cdots, x_{\Hu, M}^T)^T$, and the environment states can be represented by $x{=}(x^T_{A}, x^T_{Env})^T$ where $x_{Env}$ is the non-agent related states such as traffic lights. We use $\mathcal{I}$ to represent the social information set. For each agent, we have 
	\begin{eqnarray}
	x^{t{+}1}_{\mathcal{R}}&=&f_{\mathcal{R}}\left(x^t_{\mathcal{R}}, u^t_{\mathcal{R}}\right),\label{eq:open-loop-dynamics-R}\\
	x^{t{+}1}_{\mathcal{H}, i}&=&f_{\mathcal{H}, i}\left(x^t_{\mathcal{H}, i}, u^t_{\mathcal{H}, i}\right), i{=}1,{\cdots},M, \label{eq:open-loop-dynamics-H}
	\end{eqnarray} where $f_{\mathcal{R}}$ and $f_{\mathcal{H}, i}$ describe, respectively, the dynamics of the autonomous car and human $i$. The closed-loop dynamics of the whole multi-agent system becomes 
	\begin{equation}
	\label{eq:closed-loop-system}
	x^{t+1} = f(x^t, u^t_{\R}, u^t_{\Hu, 1}, \cdots, u^t_{\Hu, M}).
	\end{equation}
	
	We assume that all agents in the scene are noisily optimal planners. Namely, at time $t$, each agent behaves to minimize its own cost function based on its estimates of the environment states ($\hat{x}^t$) and the social information ($\hat{\mathcal{I}}^t$) inferred from observations, denoted by $o^t$. 
	Let $C_{\mathcal{R}}$ and $C_{\mathcal{H}, i}$ be, respectively, the cost functions of the robot car and human $i$ at time $t$ over a horizon of $N$: 
	\begin{eqnarray}
	&&C_{j}\left(\hat{x}^t, \hat{\mathcal{I}}^t, \mathbf{u}_{\R}, \mathbf{u}_{\Hu,i}; \theta_{j}\right)\qquad\label{eq:cost_function_general}\\
	&{=}&\sum_{k=0}^{N{-}1} c_{j}\left(\hat{x}^t, \hat{\mathcal{I}}^t, u^k_{\R}, u^k_{\Hu,i}; \theta_{j}\right), j{\in}\{\R, (\Hu,i)\}\nonumber
	\end{eqnarray}
	where $\mathbf{u}_{j}{=}(u^0_{j}, u^1_{j}, \cdots, u^{N{-}1}_{j})^T$ is the sequence of control actions of the agent within the horizon ($j{=}\R$ for the robot car and $j{=}(\Hu,i)$ for human $i$). $\theta_{j}$ with $ j{\in}\{\R, (\Hu,i)\}$ represent, respectively, the preferences of the robot car and human $i$. At every time step $t$, all agents generate their optimal sequences of actions $\mathbf{u}^*_{\R/\Hu,i}$ by minimizing their corresponding cost functions $C_{\R/\Hu,i}$, execute the first steps $u^{*0}_{\R/\Hu,i}$ (i.e., set $u^t_{\R/\Hu,i}{=}u^{*0}_{\R/\Hu,i}$ in (\ref{eq:open-loop-dynamics-R}) and (\ref{eq:open-loop-dynamics-H}) and re-plan at the next time step at $t{+}1$.
	
 As shown in (\ref{eq:cost_function_general}), robot cars generate behaviors based on their estimates of the environment states and the social information set. If the estimates significantly deviate from the ground truth, unexpected or even dangerous behaviors might be generated. For environment states, current practice is to set the estimates as the de-noised observations of the robot car from physical sensors, i.e., $\hat{x}^t{\approx}o_{\R}^t$. However, due to occlusions and limited sensor ranges, the states observation $o_{\R}^t$ of the robot car might be a subset or even different from the actual states $x^t$, which makes $\hat{x}^t{\approx}o_{\R}^t$ not an effective solution. As for the social information $\mathcal{I}^t$, it is a set of variables that cannot be directly perceived by physical sensors. Hence, to enable better autonomous driving strategies under such perception uncertainties, more advanced perception/inference scheme is desired to update $\hat{x}^t$ and $\hat{\mathcal{I}}^t$ from observations.
 
	\section{SOCIAL PERCEPTION}
	\label{sec:SP}
	Our key observation is that human traffic participants should be treated not only as dynamic obstacles that the robot cars need to be aware of, but also as distributed sensors the behaviors of which can provide additional information beyond the scope of physical sensors equipped with the autonomous vehicles.
	\subsection{Distributed Agents as Distributed Sensors}
	\label{subsec:DADS}
Again consider the multi-agent system consisting of one robot car and $M$ humans. With physical sensors subjected to occlusions and limit range, each agent can observe only a subset of the environment states, denoted by $o^t_j$ with $j{\in}\{\R,(\Hu,i)\}$. Based on the corresponding observation $o^t_j$, each agent extracts their estimates, $\hat{x}_j^t$ and $\hat{\mathcal{I}}^t_j$, on the environment states and the social information, respectively. The estimates of different agents will then influence their next-step actions/trajectories which can be perceived by other agents.
Note that due to their distributed locations, observations and the associated estimates of different agents can significantly differ, but be complementary to each other. Hence, the distributed agents can be viewed as distributed sensors which emit behavioral signals. By observing such behavioral signals, the robot car can infer estimates $\hat{x}^t_j$ and $\hat{\mathcal{I}}^t_j$ from human to reduce its perception uncertainties coming from either algorithmic limitations, or physical limitations, or both. 
	\subsection{Inference Algorithm} 
	\subsubsection{Modeling the behavior generation function of human}
	As discussed in \cref{sec:PS}, we assume that each human is a noisily optimal planner, and consider the interaction between the robot car and humans when modeling the humans' behaviors. Thus, at each time period ($N$ steps) starting at $t$, the behavior sequence $\mathbf{u}^t_{\Hu}{=}[{u}^t_{\Hu}, {u}^{t{+}1}_{\Hu}, {\cdots}, {u}^{t{+}N{-}1}_{\Hu}]^T$ minimizes the human's cost function as given in (\ref{eq:cost_function_general}) based on his/her estimates. Namely, the behavior generation function of the human can be expressed as
	\begin{IEEEeqnarray}{rCl}
		\mathbf{u}^{*,t}_{\Hu}&=&\arg\min_{\mathbf{u}_{\Hu}} C_{\Hu}\left(\hat{x}^t_{\Hu},  \hat{\mathcal{I}}_{\Hu}^t, \mathbf{u}^t_{\R}, \mathbf{u}_{\Hu}; \theta_{\Hu}\right)\quad\label{eq:behavior_generator}\\
		&\triangleq&g_{\Hu}\left(\hat{x}^t_{\Hu},  \hat{\mathcal{I}}_{\Hu}^t, \mathbf{u}^t_{\R}; \theta_{\Hu}\right),
	\end{IEEEeqnarray}
	and the optimal cost is given by 
	\begin{equation}
	\label{eq:minimum_cost}
	C^{*, t}_{\Hu}(\mathbf{u}^t_{\R})=C_{\mathcal{H}}\left(\hat{x}^t_{\Hu},  \hat{\mathcal{I}}_{\Hu}^t, \mathbf{u}^t_{\R}, \mathbf{u}^{*,t}_{\Hu}; \theta_{\mathcal{H}}\right).
	\end{equation}
	Note that in (\ref{eq:behavior_generator})~-~(\ref{eq:minimum_cost}), we model the human behavior generator $g_{\Hu}$ as his/her optimal response function to the robot car's input $\mathbf{u}^t_{\R}$ to explicitly address the influences from the robot car, as in \cite{sadighiros2016}. Hence, if the robot car can access the humans' cost function parameters $\theta_{\Hu}$ and their estimates, it can calculate the best behavioral responses $\mathbf{u}^{*,t}_{\Hu}$ from them.
	\subsubsection{Updating beliefs on estimates via inference}
To use humans as sensors of the environment, we need to construct observation models for the robot car to update its beliefs on estimates. For environment states and social information, different observation models are designed.

\vspace{5pt}
\noindent\textit{Updating beliefs on state estimates.} At every step $t$, the robot can update its beliefs on the state estimates $\hat{x}^t$ from behaviors of human $i$ via:
\begin{equation}
\label{eq: information_collection}
b_{\R}(\hat{x}^t |u_{\Hu,i}^t)\propto b_{\R}(\hat{x}^t)P(u_{\Hu,i}^t|\hat{x}^t).
\end{equation}
$P(u_{\Hu,i}^t|\hat{x}^t)$ is the probability/likelihood of human $i$ taking action $u^t_{\Hu,i}$ if the human's estimates were indeed $\hat{x}^t$. To get $P(u_{\Hu,i}^t|\hat{x}^t)$, we assume that actions with higher cost are exponentially less likely based on maximum entropy principle as in \cite{ziebart2008maximum}. This means that $P(u_{\Hu,i}^t|\hat{x}^t)$ can be approximated by:
\begin{IEEEeqnarray}{rCl}
\label{eq:single_step_IRL}
&&P(u_{\Hu,i}^t|\hat{x}^t)\propto e^{-Q^*(\hat{x}^t, \hat{\mathcal{I}}^t, \mathbf{u}^t_{\R}, u_{\Hu,i}^t; \theta_{\Hu,i})}
\end{IEEEeqnarray}
where $Q^*(\hat{x}^t, \hat{\mathcal{I}}^t, \mathbf{u}_{\R}^t, u_{\Hu,i}^t; \theta_{\R})$ represents the optimal cost to go by taking action $u^t_{\Hu,i}$, given by
\begin{IEEEeqnarray}{rCl}
	\label{eq:cost_to_go}
	&&Q^*(\hat{x}^t, \hat{\mathcal{I}}^t_i, \mathbf{u}_{\R}^t, u_{\Hu,i}^t; \theta_{\Hu,i})\nonumber\\
	&=&\min_{\mathbf{u}^{1:N{-}1}_{\Hu,i}}\sum_{k=1}^{N{-}1}c_{\Hu}\left(\hat{x}^t, \hat{\mathcal{I}}^t, u^k_{\R}, u^k_{\Hu,i}; \theta_{\Hu,i}\right).
\end{IEEEeqnarray}

\vspace{3pt}
\noindent\textit{Updating beliefs on social information.} As defined in \Cref{sec:intro}, social information refers to the group behaviors of human in the traffic scene. Therefore, to update the beliefs on estimates of social information $\hat{\mathcal{I}}$, the robot car need to collect the common behaviors from multiple human. Therefore, the belief update process becomes:
\begin{equation}
\label{eq: information_collection_social}
b_{\R}(\hat{\mathcal{I}}^t |u_{\Hu}^t)\propto b_{\R}(\hat{\mathcal{I}}^t)\prod_{i{=}1}^{M} P(u_{\Hu, i}^t|\hat{\mathcal{I}}^t),
\end{equation}
where $P(u_{\Hu, i}^t|\hat{\mathcal{I}}^t)$ is the probability/likelihood of human $i$ taking action $u_{\Hu, i}$ if the human's estimates on social information is indeed $\hat{\mathcal{I}}^t$. It can be evaluated in a similar way as (\ref{eq:single_step_IRL}) and (\ref{eq:cost_to_go}) via:
\begin{IEEEeqnarray}{rCl}
	\label{eq:single_step_IRL_social}
	&&P(u_{\Hu, i}^t|\hat{\mathcal{I}}^t)\propto e^{-Q^*(\hat{x}_t, \hat{\mathcal{I}}^t, \mathbf{u}^t_{\R}, u_{\Hu}^t; \theta_{\Hu,i})}.
\end{IEEEeqnarray}
	\subsection{Learning cost functions of human}
	As discussed above, for the robot car to update its beliefs by collecting behavioral information from  of human, the robot car needs to have access to the cost functions of human, so that it can evaluate $P(u_{\Hu}^t|\hat{x}^t)$ and $P(u_{\Hu}^t|\hat{\mathcal{I}}^t)$. One can obtain such cost functions via inverse reinforcement learning (IRL) \cite{abbeel2004apprenticeship, ziebart2008maximum, Levine2012ICML, sun2018probabilistic}. Note that during the learning process, we assume that the demonstrations are sub-optimal and there is no perception uncertainties, i.e., $\hat{x}^t{=}x^t$. A brief review of the IRL algorithm is given below.

The single-step cost is assumed to be parametrized as a linear combination of features (the social information $\mathcal{I}$ is assumed to be invariant within one horizon):
	\begin{equation}
	c(x^t, \mathcal{I}, u^t_{\R},u^t_{\Hu};\theta) = \theta^T \phi(x^t, \mathcal{I}, u^t_{\R}, u^t_{\Hu}).
	\label{eq:selfish_cost}
	\end{equation}
Over a horizon of $N$, the cumulative cost function is
	\begin{IEEEeqnarray}{rCl}
		C(x^0, \mathcal{I}, {\mathbf{u}}_{\R},{\mathbf{u}}_{\Hu};\theta)&=&\theta^T \sum_{t=0}^{N{-}1}\phi(x^t, \mathcal{I}, u^t_{\R}, u^t_{\Hu}).
	\end{IEEEeqnarray}
	Our goal is to find the weights $\theta$ which maximizes the likelihood of the demonstration set $\mathcal{U}_{D}$:
	\begin{equation}
	\theta^*=\arg\max_{\theta}P(\mathcal{U}_{D}|\theta)\label{eq:optimal_lambda}
	\end{equation}
	Building on the principle of maximum entropy, we assume that trajectories are exponentially more likely when they have lower cost:
	\begin{equation}
	{P(\mathbf{u}_{\Hu},\theta)} \propto \exp\left(-C(x^0, \mathcal{I}, \mathbf{u}_{\R},\mathbf{u}_{\Hu};\theta)\right).
	\end{equation}
	Thus the probability (likelihood) of the demonstration set becomes 
	\begin{equation}
	P(\mathcal{U}_{D}|\theta)=\Pi_{i=1}^{n}\dfrac{ P(\mathbf{u}^D_{\Hu,i},\theta)}{P(\theta)}=
	\Pi_{i=1}^{n}\dfrac{ P(\mathbf{u}^D_{\Hu,i},\theta)}{\int P(\tilde{\mathbf{u}}_{\Hu},\theta)d\tilde{\mathbf{u}}_{\Hu}}\label{eq:maximum_entropy}   
	\end{equation}
	where $n$ is the number of trajectories in $\mathcal{U}_D$.

	To tackle the partition term $\int P(\tilde{\mathbf{u}}_{\Hu},\theta)d\tilde{\mathbf{u}}_{\Hu}$ in (\ref{eq:maximum_entropy}), we approximate $C(x^0, \mathcal{I},  {\mathbf{u}}_{\R},\tilde{\mathbf{u}}_{\Hu};\theta)$ with its Laplace approximation as proposed in \cite{Levine2012ICML}:
	\begin{IEEEeqnarray}{rCl}
		&&C(x^0{,}\mathcal{I}{,}{\mathbf{u}}_{\R},\tilde{\mathbf{u}}_{\Hu};\theta)\nonumber\\
		&{\approx}&C(x^0,\mathcal{I}{,} {\mathbf{u}}_{\R},{\mathbf{u}}^D_{\Hu,i};\theta){+}\left(\tilde{\mathbf{u}}_{\Hu}{-}{\mathbf{u}}^D_{\Hu,i}\right)^{T}\dfrac{\partial C}{\partial {\mathbf{u}}_{\Hu}}\nonumber\\
		&&+\dfrac{1}{2}\left(\tilde{\mathbf{u}}_{\Hu}{-}{\mathbf{u}}^D_{\Hu,i}\right)^T\dfrac{\partial^2 C}{{\partial} \mathbf{u}^2_{\Hu}}\left(\tilde{\mathbf{u}}_{\Hu}{-}{\mathbf{u}}^D_{\Hu,i}\right).\nonumber
		\label{eq:laplace_approximation}
	\end{IEEEeqnarray}
	With the assumption of locally optimal demonstrations, we have $\dfrac{\partial C}{\partial \mathbf{u}_{\Hu}}|_{\mathbf{u}^D_{\Hu,i}}{=}0$ in (\ref{eq:laplace_approximation}). This simplifies the partition term $\int P(\tilde{\mathbf{u}}_{\Hu},\theta)d\tilde{\mathbf{u}}_{\Hu}$ as a Gaussian Integral where a closed-form solution exists (see \cite{Levine2012ICML} for details). Substituting (\ref{eq:maximum_entropy}) and (\ref{eq:laplace_approximation}) into (\ref{eq:optimal_lambda}) yields the optimal parameter $\theta^*$ as the maximizer.
	
	\section{HUMAN-LIKE BEHAVIOR PLANNING WITH SOCIAL PERCEPTION}
	In this section, we will discuss how to integrate the social perception into the decision-making and planning module to enable a more human-like driving strategy in terms of defensiveness, non-conservativeness, and social compatibility.
	\subsection{The behavior planner under uncertainties}
	\label{sec:planning}
	Due to the probabilistic nature of beliefs in (\ref{eq: information_collection}) and (\ref{eq: information_collection_social}), we utilize a probabilistic framework based on Model Predictive Control (MPC) as in \cite{zhan_non-conservatively_2016} as the planner for the autonomous cars.
	The cost function of the robot car is defined as an expected cost over the beliefs:
	\begin{IEEEeqnarray}{rCl}
		C_{\R}&{=}&\mathbb{E}_{b(\hat{x}^t), b(\hat{\mathcal{I}}^t)}C_{\R}\left(\hat{x}^t, \hat{\mathcal{I}}^t, \mathbf{u}_{\R}; \theta_{\R}\right)\qquad\label{eq:cost_function_probabilistic}\\
		&{=}&\mathbb{E}_{b(\hat{x}^t), b(\hat{\mathcal{I}}^t)}\sum_{k=0}^{N{-}1} c_{\R}\left(\hat{x}^{t,k}, \hat{\mathcal{I}}^t, u^k_{\R}; \theta_{\R}\right)\nonumber
	\end{IEEEeqnarray}
	where $C_{\R}\left(\hat{x}^t, \hat{\mathcal{I}}^t, \mathbf{u}_{\R}; \theta_{\R}\right)$ is a cumulative cost over a horizon of $N$, as defined in (\ref{eq:cost_function_general}). Note that with a long horizon $N$, discrete representation of $\hat{x}^t$ and $b(\hat{x}^t)$ is practically not feasible. In this case, we will use representative motion patterns to represent $\hat{x}^t$ as in \cite{zhan2018towards}. 
	
	\vspace{3pt}
	\noindent\textit{Cost function design.} We consider safety, efficiency, comfort, and fuel consumption in the cost. Thus, we penalize the following terms and the weight of each term can be learned via IRL as addressed in Section III-C.
	\begin{itemize}
		\item tracking error: $c_{\R, err}{=}d(x^{t, k}_{\R})$ where $d(x^{t, k}_{\R})$ is the distance from the position of the robot car at $k$ time step to the desirable traffic-free reference path.
		\item safety term: we use the relative distances from surrounding participants to evaluate the safety term. Define
		$c_{\R, safe}{=}\sum_{i=1}^{M} e^{-d_i(x^{t, k}_{\R}, x^{t, k}_{\Hu, i})}$,
		where $M$ is the number of surrounding cars and $d_i$ is the distance of the robot car to the $i$-th one. Note that $x^{t, k}_{\Hu, i}$ can be obtained via the behavior generator in (\ref{eq:behavior_generator}) and the dynamics equation in (\ref{eq:open-loop-dynamics-H}) based on current beliefs $b(x^t),  b(\mathcal{I}^t)$.
		\item efficiency: we penalize the difference between the speed of the robot car $v_{\R}{\in}x_{\R}$ and the traffic limit $v_{traffic}$, given by $c_{\R, speed}{=}(v^{t,k}_{\R}{-}v_{traffic})^2$. 
		\item acceleration: $c_{\R, acc}{=}a^{t,k}_{\R}$ where $a^{t,k}_{\R}{\in} u^{t,k}_{\R}$ is the acceleration input at $k$ time step of the robot car.
		\item jerk: $c_{\R, jerk}{=}a^{t,k}_{\R}-a^{t,k{-}1}_{\R}$.
	\end{itemize}
	Note that $v_{traffic}$ in the cost function belongs to the set of social information. We use $v_{traffic}$ instead of $v_{\lim}$ to allow the robot car to infer current traffic speed and follow it. To assure that the robot car does not break the traffic rules, we expose the maximum allowable speed limit as a constraint below.
	
	\vspace{2pt}
	\noindent\textit{Constraints.} To guarantee the feasibility of the planned trajectories, we introduce the following constraints:
	\begin{itemize}
		\item kinematics constraints: we use Bicycle model \cite{rajamani2011vehicle} to describe the kinematics model of the robot cars. 
		\item dynamics constraints: we constrain curvatures and accelerations of the vehicle as follows:
		\begin{IEEEeqnarray}{rCl}
			\vert\kappa^{t,k}_{\R}\vert \le \kappa_{\max},\quad
			 \vert a^{t,k}_{\R}\vert \le a_{\max}, k{=}0, 1, {\cdots}, N{-}1.\qquad
		\end{IEEEeqnarray}
	where $\kappa^{t,k}$ is the curvature of the planned trajectory at $k$-th step, and $\kappa_{\max}$ is the boundary of feasible curvatures. Both $\kappa_{\max}$ and $a_{\max}$ can be calculated via the ``G-G'' diagram as in \cite{rajamani2011vehicle}.
	\item safety constraints: safety constraints come from both static road structures as well as dynamic obstacles such as human drivers and pedestrians. For static structures, we use polygons to represent them, and check the robot car's distance to the polygons. For dynamic obstacles, we use several circles to cover them, and calculate distances between the robot car and the circles as in \cite{ziegler2014making}.
	\end{itemize}
Note that in the probabilistic planner, constraints over all the beliefs should all be considered. To deal with the tailing effect, we set a threshold $\epsilon$ in practice \cite{zhan_non-conservatively_2016}. This means that if the belief of a certain state or a certain social variable is lower than $\epsilon$, we will set the probability to zero in the expected cost function, and ignore the related constraints. 
	
	\subsection{The planning framework with social perception}
	With the probabilistic planner formulated in \Cref{sec:planning}, implementation of the behavior planning framework with social perception is summarized as below:
	\begin{algorithm}
		\caption{Behavior Planning with Social Perception}
		\begin{algorithmic}[1]
			\renewcommand{\algorithmicrequire}{\textbf{Input:}}
			\renewcommand{\algorithmicensure}{\textbf{Output:}}
			\REQUIRE uncertain state vector $\hat{\mathbf{x}}^{t_0}_{u}\triangleq\{\hat{x}^{t_0}_{u, 1}, \cdots, \hat{x}^{t_0}_{u, L}\}$ ($L$ is the number of possible states), observation $o^{t_0}$, prior belief $b(\hat{\mathbf{x}}_{u}^{t_0})$ and $b(\hat{\mathcal{I}}^{t_0})$, human behaviors $u^{t_0}_{\Hu, i}$
			\ENSURE  $\hat{x}^{t}$, $\hat{\mathcal{I}}^{t}$, and $u^{t}_{\R}$ for $t{>t_0}$
			\vspace{5pt}
			\\ \textbf{\textit{Initialisation}}: $\hat{\mathbf{x}}^{t_0}=\hat{\mathbf{x}}^{t_0}_{\Hu}= o^{t_0}$, 
			\FOR {$t = t_0, t_0{+}1, t_0{+}2, \cdots$}
			
			\IF {update posterior beliefs on uncertain states $x^{t}_u$,}
			\STATE 1) select human $i$ based on locations,\\
			\STATE 2) $b(\hat{x}_{u, l}^{t+1}|u_{\Hu,i}^t) \propto b(\hat{x}_{u, l}^{t})P(u_{\Hu, i}^t|\hat{x}_{u, l}^{t}), l{=}1,{\cdots}, L$\\
			\STATE 3) normalize $b(\hat{x}_{u, l}^{t})$.			
			\STATE 4) fuse new estimates of $\hat{x}_{u}^{t}$ via updated beliefs with other states to get $\hat{x}^{t}$;
			\ENDIF
			
			\IF {update posterior beliefs on social information $\mathcal{I}^{t}$,}
			\STATE 1) $b(\hat{\mathcal{I}}^{t+1}|u_{\Hu, i}^t) \propto b(\hat{\mathcal{I}}^{t})\prod_{i{=}1}^M P(u_{\Hu, i}^t|\hat{\mathcal{I}}^{t})$\\
			\STATE 2) normalize $b_{\R}(\hat{\mathcal{I}}^{t})$ and update its estimate.
			\ENDIF
			
			\STATE behavior generation via MPC: \\
			\qquad substitute $\hat{x}^{t}$, $\hat{\mathcal{I}}^{t}$ and the behavior generators for human in (\ref{eq:behavior_generator}) into the probabilistic MPC planner in \Cref{sec:planning}, and solve for the optimal actions $\mathbf{u}^t_{\R}$.
			
			\STATE execute the first action in $\mathbf{u}^t_{\R}$ as $u^t_{\R}$.
			
			\STATE update prior beliefs: $b(\hat{x}_{u, l}^{t+1})=b(\hat{x}_{u, l}^{t+1}|u_{\Hu,i}^t)$ and $b(\hat{\mathcal{I}}^{t+1})=b(\hat{\mathcal{I}}^{t+1}|u_{\Hu, i}^t)$.
			\ENDFOR
		\end{algorithmic} 
	\end{algorithm}
	
	\section{SIMULATION RESULTS}
	\label{sec:simulation}
	In this section, we give an exemplar scenario with sensor occlusions to verify the effectiveness of the proposed planning framework with social perception.
	
\begin{figure}[thpb]
		\centering
		\includegraphics[height=.2\textwidth]{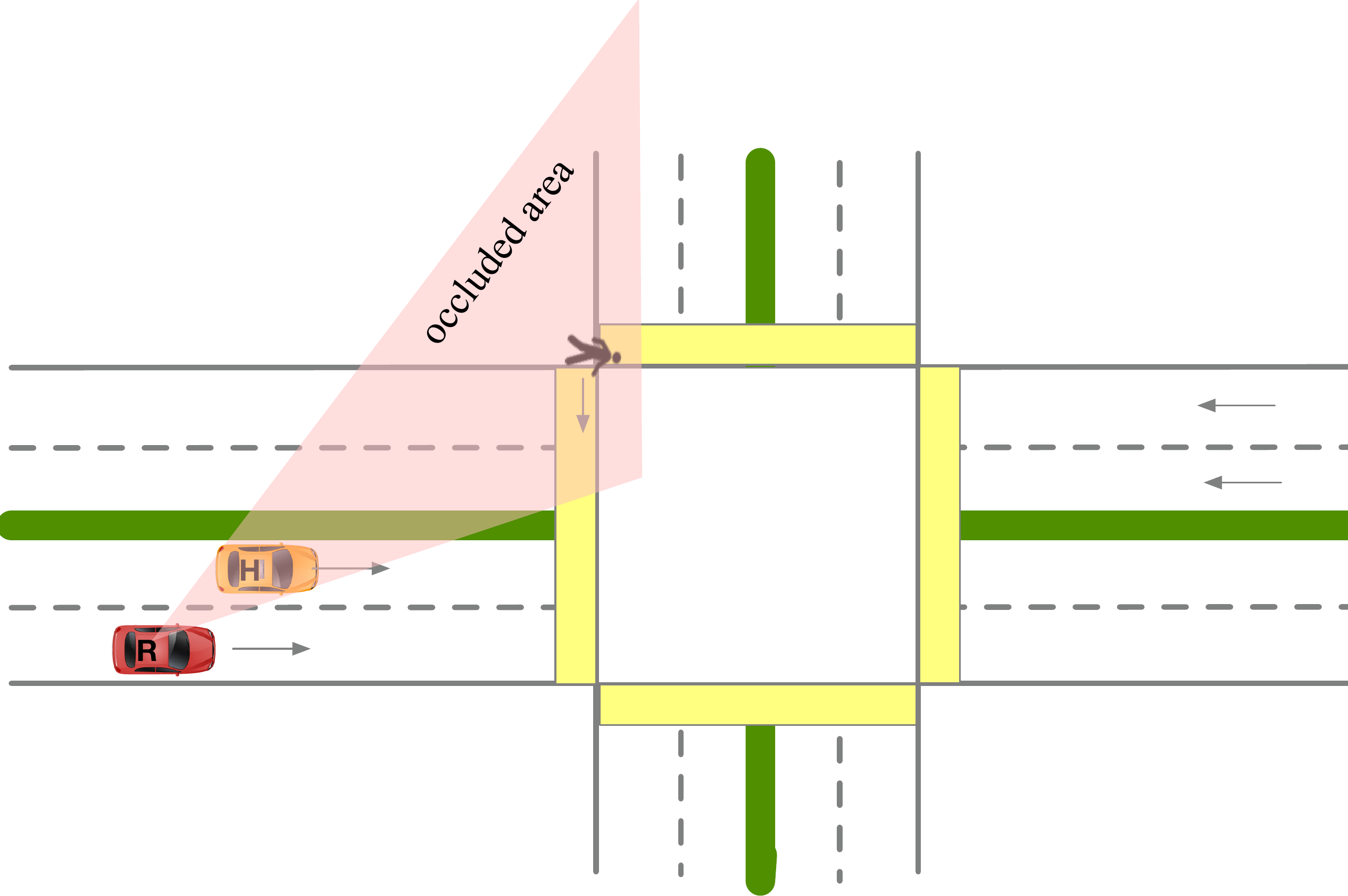}
		\caption{An example scenario of the robot car with sensor occlusions}
		\label{fig:example_1_fiture}
	\end{figure}

	Despite progresses in advanced perception and tracking algorithms, sensor occlusions are inevitable for autonomous vehicles. Consider the scenario described in Fig.~\ref{fig:example_1_fiture}, where the autonomous car (red) and a human-driven car (yellow) are driving side-by-side, and a pedestrian is about to cross the street. Due to the relative positions between the robot car and the human car, the view of the robot car is blocked by the human-driven car so that it cannot detect the pedestrian.

	In such a scenario, a conservative autonomous car will assume that there might be potential out-of-view pedestrians crossing the street. Hence, it slows down to prepare for stops or to leave a larger gap with the human driver to get better view. Both strategies will sacrifice the efficiency of the autonomous car. On the other hand, an aggressive autonomous car might directly ignore the possibility of pedestrian crossing the street and plan to drive through the intersection directly, which might lead to a collision. We note that in either case, the autonomous car perceives the environment states only via its own physical sensors, but completely ignores the information emitted via the behaviors of the human driver.
	
As shown in Fig.~\ref{fig:example_1_fiture}, the view of the human driver in this scenario is not occluded about pedestrians crossing the street, and he/she is closer to the pedestrians if there was any. Hence, from the behavior of the human driver, the robot car can actually infer and become more confident about the probability of a pedestrian crossing.

\begin{figure}[h!]
	\centering
	\includegraphics[height=.43\textwidth]{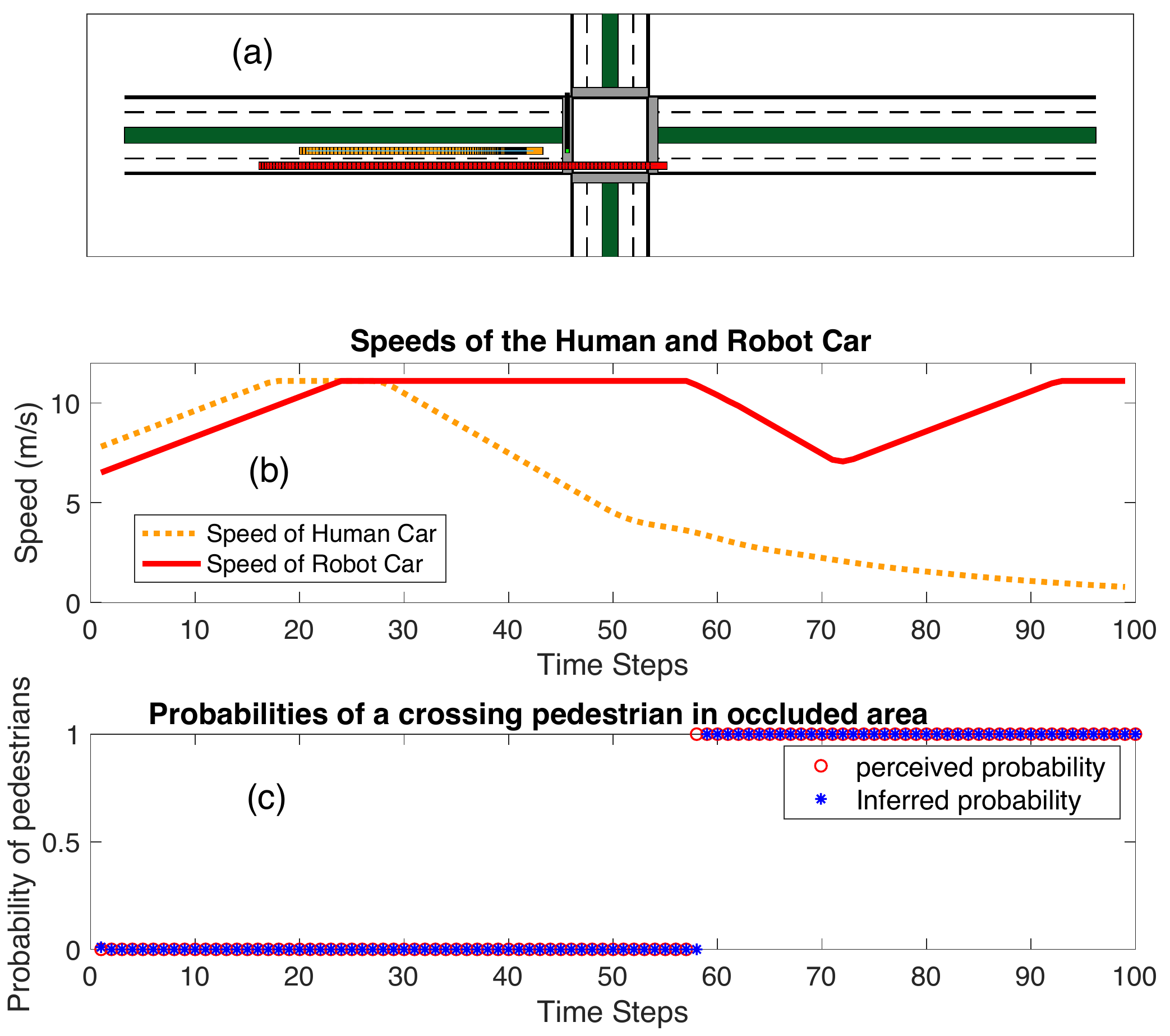}
	\caption{Simulation results with an aggressive planner with a crossing pedestrian in occluded area: (a) yellow box - human driver; red box - robot car; green dot - pedestrian; (b) speeds of the robot and human cars; (c) the perceived and inferred probabilities of pedestrians (no inference in this case)}
	\label{fig:simulation_result_1}
\end{figure}

\begin{figure}[h!]
	\centering
	\includegraphics[height=.42\textwidth]{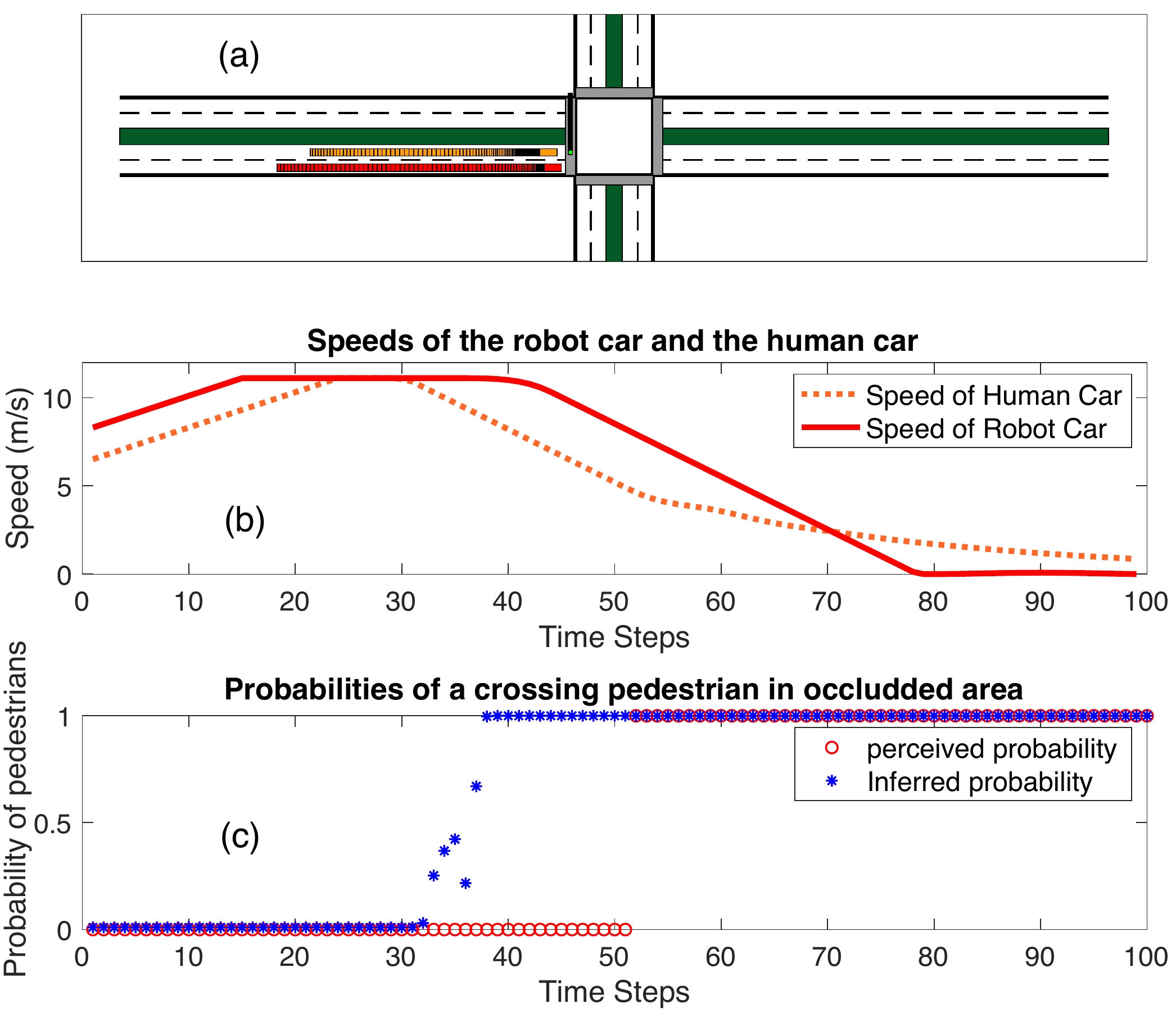}
	\caption{Simulation results with the proposed framework with a crossing pedestrian in occluded area: (a) yellow box - human driver; red box - robot car; green dot - pedestrian; (b) speeds of the robot and human cars; (c) the perceived and inferred probabilities of pedestrians}
	\label{fig:simulation_result_2}
\end{figure}

\begin{figure}[h!]
	\centering
	\includegraphics[height=.435\textwidth]{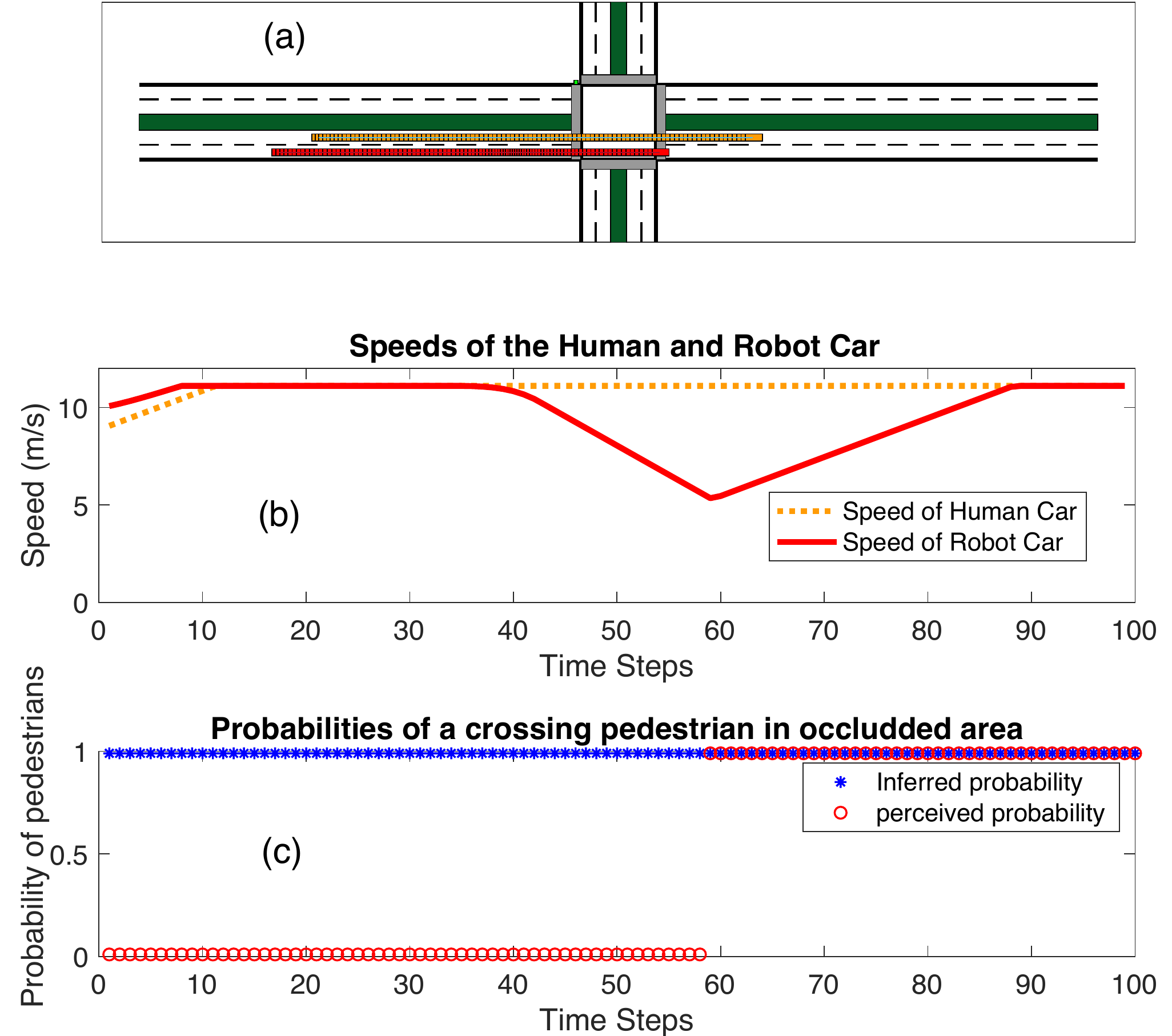}
	\caption{Simulation results with a conservative planner with non-crossing pedestrians in the occluded area: (a) yellow box - human driver; red box - robot car; green dot - pedestrian; (b) speeds of the robot and human cars; (c) the perceived and inferred probabilities of pedestrians (no inference in this case)}
	\label{fig:simulation_result_3}
\end{figure}

\begin{figure}[h!]
	\centering
	\includegraphics[height=.39\textwidth]{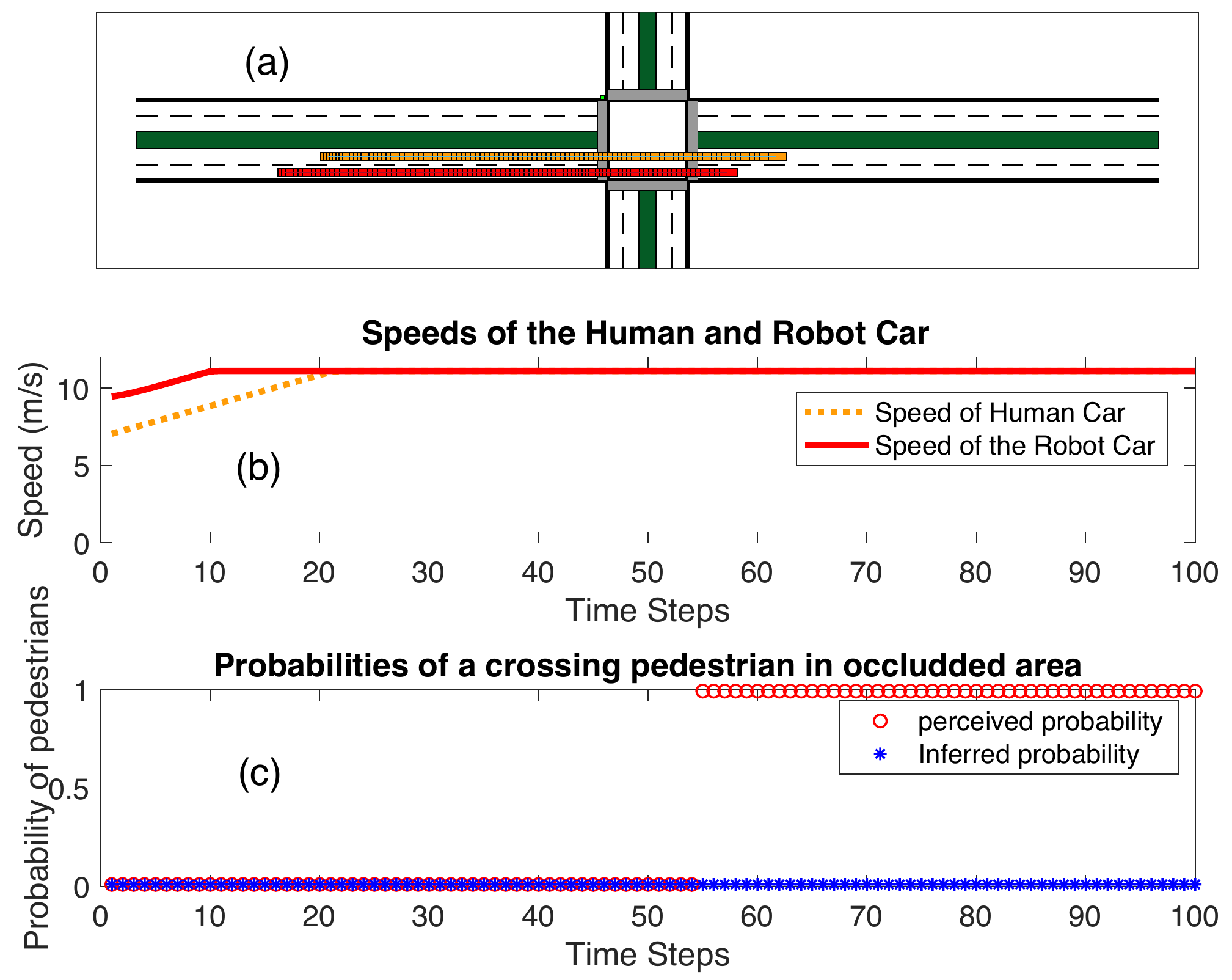}
	\caption{Simulation results with the proposed framework with non-crossing pedestrians in the occluded area: (a) yellow box - human driver; red box - robot car; green dot - pedestrian; (b) speeds of the robot and human cars; (c) the perceived and inferred probabilities of pedestrians}
	\label{fig:simulation_result_4}
\end{figure}

We simulated this traffic scenario with a conservative planner, an aggressive planner and our proposed planner with social perception. The sampling period for each time step is $0.1$s. Both cases with and without crossing pedestrians are considered, and the results are shown in Figs.~\ref{fig:simulation_result_1} through \ref{fig:simulation_result_4}. 

\noindent\textit{With crossing pedestrians in the occluded area.} Figures \ref{fig:simulation_result_1} and \ref{fig:simulation_result_2} show the comparison results with an aggressive planner and the proposed planner when there is a crossing pedestrian in the occluded area. We can see that in Fig.~\ref{fig:simulation_result_2}, with the proposed planner, when the human driver slowed down, the robot car's belief on the existence of pedestrians increased quickly (Fig.~\ref{fig:simulation_result_2}(c)). Compared to the aggressive planner in Fig.~\ref{fig:simulation_result_1}, the updated belief enables the robot car to prepare for stops before occlusions are clear, while the aggressive planner failed to yield to the pedestrian even if it braked hard when it saw the pedestrian, as in Fig.~\ref{fig:simulation_result_1}. 

\noindent\textit{With non-crossing pedestrians in the occluded area.} We also compared the proposed planner with a conservative planner which assumes the existence of crossing pedestrians in default. Results are given in Figures \ref{fig:simulation_result_3} and \ref{fig:simulation_result_4}. We can see that the proposed planner (Fig.~\ref{fig:simulation_result_4}) is much more efficient than the conservative planner (Fig.~\ref{fig:simulation_result_3}). The autonomous car with the conservative planner slowed down even if the human driver did not. On the other hand, with the proposed planner, the belief on the existence of crossing pedestrians remained low by observing the behavior of the human driver, as shown in Fig.~\ref{fig:simulation_result_4}(c), which enables the robot car to maintain relatively high speed and improves its efficiency. 



	\section{CONCLUSIONS}
	In this paper, we proposed a unified probabilistic planning framework with social perception to deal with uncertainties from physical states, prediction of others, and unknown social information. We treated all road participants as sensors in a distributed sensor network. By observing their individual behaviors as well as group behaviors, uncertainties of different types can be reduced via a uniform belief update process. We also explicitly incorporated the social perception scheme with a probabilistic planner based on MPC, which can thus generate behaviors which are defensive but not overly conservative, and socially compatible for autonomous vehicles. Simulation results in a traffic scene with sensor occlusions were given, with comparison to a conservative planner and an aggressive planner. The results showed that the proposed framework can enable more efficient and yet defensive behaviors in the presence of perception uncertainties.
	



	\bibliographystyle{IEEEtran}
	\bibliography{IV2019}
\end{document}